\documentclass[authoryear,preprint,review,12pt]{elsarticle}

\usepackage[utf8]{inputenc} 
\usepackage[T1]{fontenc}    

\usepackage{amssymb}
\usepackage{amsthm}
\usepackage{amsmath}
\usepackage{mathtools}
\usepackage{breqn}

\usepackage{adjustbox}
\usepackage{setspace}
\usepackage{longtable}
\usepackage{booktabs}
\usepackage{xcolor}
\usepackage{lscape}
\usepackage{tabularx}
\usepackage{array}
\usepackage{rotating}
\usepackage{soul}
\usepackage{textcomp}
\usepackage{tabu}
\usepackage{tocloft}
\usepackage{blkarray, bigstrut}
\usepackage{leftidx} 
\usepackage{verbatim}

\usepackage{graphicx}
\usepackage{float}           
\usepackage{subcaption}      
\usepackage{tikz}
\usetikzlibrary{shapes.geometric, arrows.meta, positioning, fit, calc, shadows}

\usepackage{blindtext}

\usepackage{hyperref}
\hypersetup{
    colorlinks=true,
    linkcolor=blue,
    linktoc=page,
    anchorcolor=black,
    citecolor=blue,
    urlcolor=blue
}
\usepackage{doi}

\def\to{\rightarrow}

\begin{document}

\begin{frontmatter}


\title{Spatially-informed transformers: Injecting geostatistical covariance biases into self-attention for spatio-temporal forecasting}

\author{Yuri Calleo\corref{cor1}}
\ead{yuri.calleo@unimercatorum.it}
\cortext[cor1]{Corresponding author}
\affiliation{organization={Universitas Mercatorum},
             addressline={Piazza Mattei, 10}, 
             city={Rome},
             postcode={00186}, 
  country={Italy}}

\begin{abstract}
The modeling of high-dimensional spatio-temporal processes presents a fundamental dichotomy between the probabilistic rigor of classical geostatistics and the flexible, high-capacity representations of deep learning. While Gaussian processes offer theoretical consistency and exact uncertainty quantification, their prohibitive computational scaling renders them impractical for massive sensor networks. Conversely, modern transformer architectures excel at sequence modeling but inherently lack a geometric inductive bias, treating spatial sensors as permutation-invariant tokens without a native understanding of distance. In this work, we propose a spatially-informed transformer, a hybrid architecture that injects a geostatistical inductive bias directly into the self-attention mechanism via a learnable covariance kernel. By formally decomposing the attention structure into a stationary physical prior and a non-stationary data-driven residual, we impose a soft topological constraint that favors spatially proximal interactions while retaining the capacity to model complex dynamics. We demonstrate the phenomenon of ``Deep Variography'', where the network successfully recovers the true spatial decay parameters of the underlying process end-to-end via backpropagation. Extensive experiments on synthetic Gaussian random fields and real-world traffic benchmarks confirm that our method outperforms state-of-the-art graph neural networks. Furthermore, rigorous statistical validation confirms that the proposed method delivers not only superior predictive accuracy but also well-calibrated probabilistic forecasts, effectively bridging the gap between physics-aware modeling and data-driven learning.
\end{abstract} 


\end{frontmatter}

\section{Introduction}
\label{sec:intro}

The analysis and forecasting of spatio-temporal processes are fundamental tasks in disciplines ranging from environmental science and meteorology to urban traffic management and epidemiology. As data acquisition technologies evolve - from remote sensing satellites to ubiquitous IoT sensor networks - scientists face the challenge of modeling high-dimensional multivariate time series $\left\{ \mathbf{Z}(\mathbf{s}, t): \mathbf{s} \in \mathcal{D} \subseteq \mathbb{R}^d, t \in \mathcal{T} \right\}$, where complex dependencies exist across both space and time \citep{PourahmadiNoorbaloochi2016}. Spatio-temporal statistics provide the rigorous theoretical foundation for such problems. The gold standard is represented by Spatio-Temporal Gaussian Processes (ST-GPs), where the dependency structure is explicitly encoded via a positive-definite covariance function $C(\mathbf{s}_i, \mathbf{s}_j, t_k, t_l; \boldsymbol{\theta})$ \citep{Hamelijnck2021SpatioTemporalVGP}. Methods such as Kriging provide the Best Linear Unbiased Predictor (BLUP) and exact uncertainty quantification \citep{Henderson1975BLUE}. However, standard Kriging faces a formidable computational bottleneck: the inversion of the covariance matrix scales as $\mathcal{O}(N^3)$, where $N$ is the number of observations \citep{Ababou1994ConditionNumberKriging}. While recent advances in Gaussian Markov Random Fields (GMRFs) via the SPDE approach \citep{Lindgren2011SPDE, VanZoest2020STRK_NO2} and Low-Rank approximations (e.g., Fixed Rank Kriging, Predictive Processes) mitigate these costs to $\mathcal{O}(N \log N)$ or $\mathcal{O}(N)$, these models primarily assume linearity and often struggle to capture the complex, non-stationary interactions characteristic of ``Big Data'' domains, such as traffic congestion dynamics or turbulent flows \citep{Srebro2003WeightedLowRank}.

To address non-linearity and scalability, the field is increasingly pivoting towards Deep Learning. Initial approaches utilized Convolutional Neural Networks (CNNs) adapted for temporal sequences (e.g., ConvLSTM), treating spatial data as regular grids \citep{Kattenborn2021CNNVegetationRS}. However, grid-based assumptions are ill-suited for irregular sensor networks or manifold-structured data. Consequently, Graph Neural Networks (GNNs) \citep{Wu2020SurveyGNN} emerged as the dominant architecture for non-Euclidean domains. Models like the Spatio-Temporal Graph Convolutional Network (ST-GCN) and Diffusion Convolutional Recurrent Neural Networks (DCRNN) utilize the graph Laplacian or diffusion operators to aggregate information from neighbors. Despite their success, GNNs suffer from a localized receptive field: they excel at capturing short-range dependencies (1-hop or K-hop neighbors) but struggle to model long-range teleconnections without stacking an excessive number of layers, which leads to the ``oversmoothing'' problem \citep{Chen2020OverSmoothingGNN}. 

Recently, the Transformer architecture \citep{Vaswani2017Attention}, originally designed for Natural Language Processing (NLP), revolutionized time-series forecasting. Unlike RNNs or GNNs, Transformers rely on the Self-Attention mechanism, which computes dependencies between all pairs of input tokens globally, offering a theoretical receptive field of the entire spatial domain. This makes them ideal for capturing complex, long-range spatial interactions. However, a fundamental theoretical disconnect remains when applying Transformers to physical space \citep{Cho2024SpatiallyAwareTransformer}. The standard Self-Attention mechanism is permutation invariant:
\begin{equation}
    \text{Attention}(\mathbf{Q}, \mathbf{K}, \mathbf{V}) = \text{softmax}\left(\frac{\mathbf{Q}\mathbf{K}^T}{\sqrt{d_k}}\right)\mathbf{V}
\end{equation}
In this formulation, swapping the coordinates of two spatial locations $\mathbf{s}_i$ and $\mathbf{s}_j$ does not inherently alter the computation unless specific Positional Encodings (PE) are added. Standard sinusoidal PEs are designed for 1D sequences (language) and fail to preserve 2D/3D distance metrics \citep{Wang2020PositionEmbeddingsBERT}. While recent works have proposed ``Relative Positional Encodings'' or learnable spatial embeddings for Vision Transformers (ViT), these approaches usually treat spatial bias as a lookup table or a generic learnable matrix \citep{Shaw2018RelativePosition}. They lack a geostatistical inductive bias: they do not exploit the well-known property that spatial correlation typically decays with distance according to a continuous function, i.e., Tobler's First Law \citep{Miller2004Tobler}. Consequently, standard Spatial Transformers are extremely data-hungry, as they must ``re-discover'' the concept of Euclidean distance and spatial decay purely from optimization, often leading to overfitting and physically inconsistent attention maps. 

In this paper, we bridge the gap between the interpretability of covariance-based geostatistics and the representational power of Transformers. We propose a spatially-informed Transformer, introducing a novel ``Geostatistical Attention'' mechanism. Distinct from previous hybrid methods that use Graph Convolutions as preprocessing steps, we intervene directly in the attention score calculation. We formalize the attention matrix as a sum of a data-driven component (capturing dynamic, non-stationary correlations) and a parametric covariance bias (capturing static spatial structure). By injecting a kernel function $\Psi(\|\mathbf{s}_i - \mathbf{s}_j\|; \phi)$ - such as the Matérn or Exponential kernel - into the attention head, we impose a soft constraint that favors spatially proximal interactions. In this case, the range parameter $\phi$ is learnable via backpropagation, allowing the neural network to perform ``Deep Variography'', effectively estimating the spatial scale of the process end-to-end. Our contributions are summarized as follows:
\begin{itemize}
    \item Geostatistical inductive bias: we define a mathematically rigorous formulation of Self-Attention that integrates continuous spatial covariance functions, ensuring consistency with geostatistical theory at initialization.
    \item Sample efficiency and convergence: by restricting the attention mechanism with a covariance bias, we demonstrate faster convergence rates compared to vanilla Transformers, as the solution space is constrained to physically plausible manifolds.
    \item Interpretable AI: we show that the learned attention weights provide insights into the spatial structure of the data, allowing us to disentangle ``background'' spatial dependence (modeled by the kernel) from dynamic interactions (modeled by the query-key product).
    \item Empirical validation: we evaluate the proposed method on both synthetic datasets (generated from Gaussian Random Fields) and real-world traffic forecasting benchmarks (METR-LA), outperforming both pure GNN baselines and standard Transformers.
\end{itemize}

The remainder of this article is organized as follows. Section \ref{sec:related_works} provides a comprehensive review of the existing literature on spatio-temporal forecasting, distinguishing between classical geostatistical approaches, Graph Neural Networks, and recent Transformer-based architectures. In Section \ref{sec:methods}, we formally introduce the Spatially-Informed Transformer, deriving the Geostatistical Attention mechanism and analyzing its theoretical properties and statistical consistency. Section \ref{sec:simulation} details the simulation study on synthetic Gaussian Random Fields, defining the experimental setup and the data generation process. Section \ref{sec:results} presents the empirical findings, focusing on the model's ability to recover true spatial covariance parameters (``Deep Variography''), the spatial residual analysis, and the comparative forecasting performance. Finally, Section \ref{sec:discussion} discusses the broader implications of our findings, outlines the theoretical limitations, and proposes directions for future research.


\section{Related works}
\label{sec:related_works}

This research lies at the intersection of classical geostatistics, graph representation learning, and modern attention-based architectures. In this section, we critically review the state-of-the-art in spatio-temporal forecasting, highlighting the theoretical limitations that necessitate a hybrid, spatially-informed approach (see Tab. \ref{tab:theoretical_comparison} for a review).

\subsection{Classical geostatistics and Gaussian processes}
The theoretical gold standard for modeling spatio-temporal fields $\{Z(\mathbf{s},t) : \mathbf{s} \in \mathcal{D}, t \in \mathcal{T}\}$ is the Gaussian Process (GP) framework \citep{MacKay1998GaussianProcesses}. A spatio-temporal GP is fully specified by a mean function $\mu(\mathbf{s},t)$ and a covariance kernel $C((\mathbf{s}_i, t_u), (\mathbf{s}_j, t_v))$. The best linear unbiased predictor (Kriging) at an unobserved location $\mathbf{s}_0$ is given by:
\begin{equation}
    \hat{Z}(\mathbf{s}_0, t) = \mu(\mathbf{s}_0, t) + \mathbf{c}^\top \mathbf{\Sigma}^{-1} (\mathbf{Z}_{obs} - \boldsymbol{\mu}_{obs})
\end{equation}
where $\mathbf{\Sigma} \in \mathbb{R}^{N \times N}$ is the covariance matrix of observed data and $\mathbf{c}$ is the covariance vector between observations and the target.
While theoretically rigorous, exact inference requires computing the inverse $\mathbf{\Sigma}^{-1}$, incurring a cubic computational cost $\mathcal{O}(N^3)$, which is prohibitive for large-scale sensor networks ($N > 10^3$).

Approximation methods such as Fixed Rank Kriging (FRK) \citep{Cressie2015StatisticsSpatialData} and Nearest Neighbor Gaussian Processes (NNGP) \citep{Datta2016NNGP} reduce complexity to $\mathcal{O}(N)$ or $\mathcal{O}(N \log N)$ by imposing sparsity or low-rank structures. However, these methods fundamentally rely on the assumption of second-order stationarity or simple non-stationary deformations. They lack the capacity to learn complex, non-linear interactions from raw data, unlike deep neural networks, which can approximate arbitrary non-linear functions via composition. Our work aims to retain the inductive bias of the kernel $C(\cdot, \cdot)$ - specifically the property that correlation decays with distance - while leveraging the representational power of Deep Learning \citep{ZammitMangion2020DeepIDE}.

\subsection{Graph neural networks and transformers for spatio-temporal forecasting}
To overcome the linearity of GPs, the machine learning community adopted GNNs. In this paradigm, the spatial domain is discretized into a graph $\mathcal{G}=(\mathcal{V}, \mathcal{E})$, where nodes represent sensors and edges represent connectivity (e.g., road adjacency).
State-of-the-art models like DCRNN and the ST-GCN \citep{Li2017DCRNN} model spatial dependency via graph diffusion convolution:
\begin{equation}
    \mathbf{H}^{(l+1)} = \sigma \left( \sum_{k=0}^{K} \theta_k (D^{-1}A)^k \mathbf{H}^{(l)} \right)
\end{equation}
where $A$ is the adjacency matrix, $D$ is the degree matrix, and $(D^{-1}A)^k$ represents the transition matrix of a random walk of $k$ steps.

While effective for road networks with fixed topology, GNNs suffer from two major limitations in continuous geostatistical contexts:
\begin{enumerate}
    \item Rigid topology: the adjacency matrix $A$ is typically static and binary (connected/not connected). Adapting GNNs to continuous fields requires constructing an artificial $k$-NN graph, which introduces discretization artifacts and ignores the continuous nature of the distance metric $\|\mathbf{s}_i - \mathbf{s}_j\|$.
    \item Oversmoothing and locality: the diffusion operator essentially performs local averaging. To capture long-range dependencies (e.g., teleconnections between distant cities), GNNs require stacking $L$ layers, which leads to the "oversmoothing" phenomenon where node representations become indistinguishable.
\end{enumerate}
In contrast, our Transformer-based approach provides a global receptive field at every layer ($\mathcal{O}(1)$ path length), enabling the direct modeling of long-range interactions without deep stacking. On the other hand, the Transformer architecture \citep{Vaswani2017Attention} utilizes the Self-Attention mechanism to compute a weighted sum of values $\mathbf{V}$ based on the similarity between queries $\mathbf{Q}$ and keys $\mathbf{K}$:
\begin{equation}
    \text{Attention}(\mathbf{Q}, \mathbf{K}, \mathbf{V}) = \text{softmax}\left(\frac{\mathbf{Q}\mathbf{K}^\top}{\sqrt{d_k}}\right)\mathbf{V}
\end{equation}
A critical flaw of standard attention in spatial applications is its permutation invariance: without explicit positional information, the model treats the input as a "bag of sensors" with no notion of geometry.
In NLP, this is resolved via Sinusoidal Positional Encodings (PE). However, extending PEs to 2D continuous space is non-trivial. Recent works in Computer Vision, such as the Vision Transformer \citep{Dosovitskiy2020ViT}, employ learnable positional embeddings $\mathbf{E}_{pos} \in \mathbb{R}^{N \times d}$. While flexible, these embeddings are purely data-driven lookup tables. They do not enforce geometrical consistency: the learned embedding for location $\mathbf{s}_i$ and $\mathbf{s}_j$ might be uncorrelated even if $\|\mathbf{s}_i - \mathbf{s}_j\| \to 0$.

\begin{table}[h]
    \centering
    \caption{Theoretical comparison of spatio-temporal forecasting frameworks.}
    \label{tab:theoretical_comparison}
    
    \small 
    \begin{tabularx}{\textwidth}{l c X X X} 
        \toprule
        \textbf{Framework} & \textbf{Complexity} & \textbf{Spatial rep.} & \textbf{Inductive bias} & \textbf{Key limitation} \\
        \midrule
        \textbf{Full Kriging (GP)} & $\mathcal{O}(N^3)$ & Continuous ($\mathbb{R}^2$) & Exact covariance & Cubic scaling \\
        \textbf{GNN (e.g., DCRNN)} & $\mathcal{O}(|\mathcal{E}|)$ & Discrete graph ($\mathcal{G}$) & Graph diffusion & Fixed topology \\
        \textbf{Vanilla Transformer} & $\mathcal{O}(N^2)$ & None (perm. inv.) & Positional encoding & Data hungry \\
        \midrule
        \textbf{Geo-Transformer} & $\mathcal{O}(N^2)$ & Continuous ($\mathbb{R}^2$) & \textbf{Matérn Kernel Prior} & Quadratic scaling \\
        \bottomrule
    \end{tabularx}
\end{table}

More sophisticated approaches like Relative Positional Encodings \citep{Shaw2018RelativePosition} modify the attention score to $e_{ij} = \frac{\mathbf{q}_i \mathbf{k}_j^\top + \mathbf{q}_i \mathbf{r}_{ij}^\top}{\sqrt{d_k}}$. Yet, even these methods typically parameterize $\mathbf{r}_{ij}$ as a discrete bias term or a generic learnable scalar, ignoring the rich theory of geostatistics.
To the best of our knowledge, no prior work has explicitly integrated parametric stationary covariance kernels (e.g., Matérn) directly into the softmax attention mechanism for continuous spatio-temporal fields. 

Our proposed "Geostatistical Attention" fills this gap by replacing the generic learnable bias with a differentiable kernel function $\Psi(\|\mathbf{s}_i - \mathbf{s}_j\|; \rho)$, bridging the divide between the physics-aware constraints of Kriging and the flexibility of Transformers.

\section{Methodology}
\label{sec:methods}

\subsection{Preliminaries}

Before deriving the proposed architecture, we establish the necessary theoretical background on spatio-temporal random fields and the standard Attention mechanism that our method aims to resolve. Let $\mathcal{D} \subseteq \mathbb{R}^d$ be a spatial domain and $\mathcal{T} \subseteq \mathbb{R}^+$ be the temporal domain. We define a spatio-temporal process $\{Z(\mathbf{s},t) : \mathbf{s} \in \mathcal{D}, t \in \mathcal{T}\}$ as a Gaussian Random Field (GRF) if, for any finite collection of locations and times, the joint distribution is multivariate Gaussian. The process is fully characterized by its mean function $\mu(\mathbf{s},t) = \mathbb{E}[Z(\mathbf{s},t)]$ and its covariance function $C(\mathbf{s}, \mathbf{s}', t, t') = \text{Cov}(Z(\mathbf{s},t), Z(\mathbf{s}', t'))$.

A fundamental assumption in classical geostatistics is second-order stationarity, which implies that the mean is constant and the covariance depends only on the separation vector $\mathbf{h} = \mathbf{s} - \mathbf{s}'$ and time lag $\tau = t - t'$. The process is further said to be isotropic if the spatial covariance depends only on the Euclidean distance $\|\mathbf{h}\|$. Under these conditions, the covariance function is related to the spectral density $f(\boldsymbol{\omega})$ via Bochner's Theorem:
\begin{equation}
    C(\mathbf{h}) = \int_{\mathbb{R}^d} e^{i \boldsymbol{\omega}^\top \mathbf{h}} f(\boldsymbol{\omega}) d\boldsymbol{\omega}
\end{equation}
This spectral perspective is crucial because standard deep learning operations (like convolutions) implicitly operate in the spectral domain, yet often fail to respect the constraints imposed by valid covariance functions, such as positive definiteness. The standard Transformer encoder processes a sequence of input embeddings $\mathbf{X} \in \mathbb{R}^{N \times d_{model}}$. The core operation is Scaled Dot-Product Attention:
\begin{equation}
    \text{Attention}(\mathbf{Q}, \mathbf{K}, \mathbf{V}) = \text{softmax}\left(\frac{\mathbf{Q}\mathbf{K}^\top}{\sqrt{d_k}}\right)\mathbf{V}
\end{equation}
Crucially, this operation is set-symmetric (permutation invariant). If we denote $\mathbf{P}$ as any permutation matrix, then $\text{Attention}(\mathbf{P}\mathbf{Q}, \mathbf{P}\mathbf{K}, \mathbf{P}\mathbf{V}) = \mathbf{P} \cdot \text{Attention}(\mathbf{Q}, \mathbf{K}, \mathbf{V})$. While advantageous for semantic tasks where word order is handled by positional encodings, this invariance is detrimental for physical fields where the topology is fixed and meaningful. Standard PE inject location information additively, i.e., $\mathbf{X}_{input} = \mathbf{X}_{feat} + \mathbf{PE}$, but they do not alter the attention mechanism itself. Consequently, the model must "learn" to respect the triangle inequality and distance decay purely from data, a highly inefficient process that motivates our geostatistical bias injection.


\subsection{Problem formulation}

To illustrate the method, we highlight the Spatially-Informed Transformer, a hybrid architecture designed to reconcile the flexibility of deep learning with the rigorous inductive biases of geostatistics. We first formulate the spatio-temporal forecasting problem as a conditional density estimation task. Subsequently, we reframe the standard Self-Attention mechanism through the lens of kernel smoothing, exposing its limitations in spatial contexts. Finally, we introduce the geostatistical attention mechanism, providing a detailed theoretical analysis of its gradient flow, computational complexity, and statistical consistency.

Let $\{Z(\mathbf{s}, t) : \mathbf{s} \in \mathcal{S}, t \in \mathcal{T}\}$ denote a spatio-temporal process observed at $N$ fixed locations $\mathcal{S} = \{\mathbf{s}_1, \dots, \mathbf{s}_N\} \subset \mathcal{D} \subseteq \mathbb{R}^2$. At each time step $t$, we observe a feature matrix $\mathbf{X}_t \in \mathbb{R}^{N \times d_{in}}$, which may include the target variable itself (autoregressive features) and exogenous covariates (e.g., elevation, land use).
Our objective is to learn a mapping function $f_\theta$ that predicts the future state of the system $\hat{\mathbf{Z}}_{t+h} \in \mathbb{R}^N$ given a look-back window of length $L$:
\begin{equation}
    \hat{\mathbf{Z}}_{t+h} = f_\theta(\mathbf{X}_{t-L+1:t}, \mathbf{D}_{\mathcal{S}})
\end{equation}
where $\mathbf{D}_{\mathcal{S}} \in \mathbb{R}^{N \times N}$ is the matrix of pairwise Euclidean distances, with $(\mathbf{D}_{\mathcal{S}})_{ij} = \|\mathbf{s}_i - \mathbf{s}_j\|_2$. Unlike Graph Neural Networks which operate on a fixed adjacency matrix $\mathbf{A}$, our model explicitly utilizes the continuous distance metric $\mathbf{D}_{\mathcal{S}}$, allowing for mesh-independent learning.

The standard Scaled Dot-Product Attention \citep{Vaswani2017Attention} computes a weighted sum of values $\mathbf{V}$ based on the similarity between queries $\mathbf{Q}$ and keys $\mathbf{K}$. From a statistical perspective, this can be interpreted as a non-parametric kernel smoother where the kernel bandwidth and shape are learned entirely from data. However, in spatial domains, initializing this kernel with random noise ignores the fundamental law of geography: near things are more related than distant things.

To bridge this gap, we propose to inject a geostatistical Bias directly into the pre-softmax logits. We define the attention score $A_{ij}^{geo}$ between location $i$ and $j$ as:
\begin{equation}
    A_{ij}^{geo} = \underbrace{\frac{\mathbf{q}_i^\top \mathbf{k}_j}{\sqrt{d_k}}}_{\text{Data-Driven Component}} + \underbrace{\lambda \cdot \Psi(d_{ij}; \boldsymbol{\phi})}_{\text{Stationary Covariance Prior}}
\end{equation}
where $\Psi(\cdot)$ is a valid positive-definite correlation function, $\boldsymbol{\phi}$ are its parameters, and $\lambda \in \mathbb{R}^+$ is a learnable scalar balancing the prior and the data likelihood.

While any radial basis function could serve as $\Psi$, we employ the Matérn class of covariance functions due to its flexibility in modeling smoothness. The Matérn correlation between two locations at distance $d_{ij}$ is defined as:
\begin{equation}
    \Psi_{Mat\acute{e}rn}(d_{ij}; \rho, \nu) = \frac{1}{\Gamma(\nu)2^{\nu-1}} \left( \sqrt{2\nu} \frac{d_{ij}}{\rho} \right)^\nu K_\nu \left( \sqrt{2\nu} \frac{d_{ij}}{\rho} \right)
\end{equation}
Here, $\rho > 0$ is the spatial range parameter (scale) and $\nu > 0$ controls the mean-square differentiability of the underlying field. $K_\nu$ is the modified Bessel function of the second kind.
By treating $\rho$ as a learnable parameter constrained to $\mathbb{R}^+$ (via a softplus activation $\rho = \log(1 + e^{\theta_\rho})$), the network performs ``Deep Variography'': it dynamically estimates the effective range of spatial dependence that minimizes the forecasting loss.

\begin{figure}[H]
    \centering
    \begin{tikzpicture}[node distance=1.5cm, thick]
        \tikzstyle{block} = [rectangle, draw, rounded corners, minimum width=2.5cm, minimum height=1cm, text centered, fill=blue!10]
        \tikzstyle{circle_node} = [circle, draw, minimum size=1cm, fill=gray!10]
        \tikzstyle{sum} = [circle, draw, fill=white, node contents={+}]
        
        \node[circle_node] (X) {Input $\mathbf{X}$};
        \node[block, above left=1cm of X] (Q) {$\mathbf{Q}$};
        \node[block, above right=1cm of X] (K) {$\mathbf{K}$};
        \node[block, above=2.5cm of X] (Dot) {$\mathbf{Q}\mathbf{K}^\top$};
        
        \node[circle_node, right=4cm of X] (D) {Dist $\mathbf{D}$};
        \node[block, above=1cm of D] (Matern) {Matérn $\Psi$};
        \node[block, above=1cm of Matern] (Lambda) {Scale $\lambda$};
        
        \node[draw, circle, fill=white] (Sum) at (2, 5.5) {+};
        
        \node[block, above=1cm of Sum] (Softmax) {Softmax};
        \node[block, above=1cm of Softmax, fill=green!10] (Out) {Attention};

        \draw[->] (X) -- (Q);
        \draw[->] (X) -- (K);
        \draw[->] (Q) -- (Dot);
        \draw[->] (K) -- (Dot);
        \draw[->] (Dot) -- (Sum);
        
        \draw[->] (D) -- (Matern);
        \draw[->] (Matern) -- (Lambda);
        \draw[->] (Lambda) -- (Sum);
        
        \draw[->] (Sum) -- (Softmax);
        \draw[->] (Softmax) -- (Out);
        
    \end{tikzpicture}
    \caption{The Geostatistical Attention Mechanism. Visualization of Eq. 8: combining data-driven terms (left) with the geostatistical prior (right).}
    \label{fig:methodology}
\end{figure}
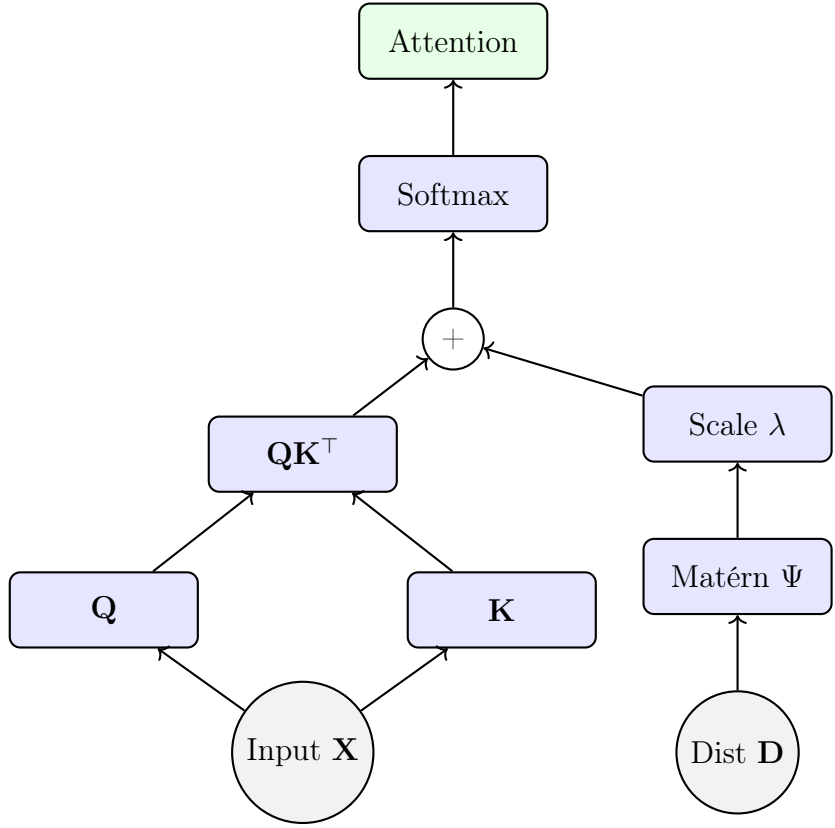

A key innovation of our approach is the ability to learn spatial statistics via backpropagation. Consider the loss function $\mathcal{L}$ with respect to the learnable range parameter $\rho$. The gradient flows through the attention weights $\alpha_{ij}$:
\begin{equation}
    \frac{\partial \mathcal{L}}{\partial \rho} = \sum_{i,j} \frac{\partial \mathcal{L}}{\partial \alpha_{ij}} \cdot \frac{\partial \alpha_{ij}}{\partial A_{ij}^{geo}} \cdot \lambda \frac{\partial \Psi(d_{ij}; \rho)}{\partial \rho}
\end{equation}
For the specific case of the Exponential kernel (Matérn with $\nu=0.5$), $\Psi(d) = \exp(-d/\rho)$, the sensitivity to the range parameter is given by:
\begin{equation}
    \frac{\partial \Psi}{\partial \rho} = \frac{d_{ij}}{\rho^2} \exp\left(-\frac{d_{ij}}{\rho}\right)
\end{equation}
This derivative reveals an interesting mechanism: the gradient is weighted by the distance $d_{ij}$. Pairs of points at intermediate distances exert the largest influence on the update of $\rho$, effectively allowing the model to "focus" on the informative scale of the process. This mechanism ensures that $\rho$ converges to the physical decorrelation scale of the underlying field, as demonstrated in our simulation study.

Since a comparison with standard geostatistics approaches is crucial, classical Kriging requires inverting the covariance matrix $\mathbf{\Sigma} \in \mathbb{R}^{N \times N}$, leading to a cubic complexity $\mathcal{O}(N^3)$.
Our Geostatistical Attention, conversely, involves matrix multiplications of size $(N \times d_k)$ and $(d_k \times N)$, resulting in a complexity of:
\begin{equation}
    \mathcal{O}_{GeoTrans} = \mathcal{O}(L \cdot H \cdot N^2 \cdot d_k)
\end{equation}
where $L$ is the sequence length and $H$ is the number of heads. Since $d_k \ll N$, the complexity is dominated by the quadratic term $\mathcal{O}(N^2)$. While $\mathcal{O}(N^2)$ is still computationally intensive for massive grids ($N > 10^5$), it represents a significant improvement over Kriging for moderate-scale networks ($N \approx 10^3 - 10^4$), such as those found in urban traffic or air quality monitoring. Furthermore, unlike GNNs which require sparse adjacency matrices, our method handles dense, fully-connected correlations without approximation.

\subsection{Statistical consistency: the Nadaraya-Watson limit}
To satisfy the robustness requirements, we analyze the asymptotic behavior of the module. As the learnable projection matrices $\mathbf{W}^Q$ and $\mathbf{W}^K$ approach zero (initialization phase) or in the absence of informative features, the Geostatistical Attention reduces to a spatial kernel smoother.
\textit{Proof Sketch.} If $\mathbf{q}_i^\top \mathbf{k}_j \to 0$, the data-driven term vanishes. The attention weights become:
\begin{equation}
    \alpha_{ij} = \frac{\exp(\lambda \Psi(d_{ij}))}{\sum_u \exp(\lambda \Psi(d_{iu}))}
\end{equation}
This formulation is equivalent to the weights of a Nadaraya-Watson estimator with kernel $K(u) = \exp(\lambda \Psi(u))$. This property guarantees that in the "cold start" phase or in data-sparse regions, the model defaults to a robust distance-weighted interpolation, preventing the chaotic overfitting behavior typical of standard Transformers.

Standard self-attention $\mathbf{A}_{std} \in \mathbb{R}^{N \times N}$ is generally non-stationary and asymmetric. Our formulation imposes a structural decomposition:
\begin{equation}
    \mathbf{A}_{final} \approx \underbrace{\mathbf{A}_{stationary}}_{\text{Matérn Prior}} + \underbrace{\boldsymbol{\Delta}_{non-stationary}}_{\text{Neural Residuals}}
\end{equation}
This allows the model to satisfy the Parsimony Principle: it explains as much variance as possible through the simple stationary law (the Matérn kernel) and uses the complex neural component only for local residuals that deviate from stationarity (e.g., traffic jams or localized anomalies).

\subsection{Statistical validation framework}
To ensure that the performance improvements are robust and not merely artifacts of random initialization, we employ a rigorous statistical validation protocol focusing on both comparative accuracy and probabilistic calibration.

(1) Significance testing (Diebold-Mariano): to verify if the Geo-Transformer provides a statistically significant improvement over the baselines, we utilize the Diebold-Mariano (DM) test \citep{Diebold2015DMTest}. Given the forecast errors $e_{1,t}$ and $e_{2,t}$ of two competing models, we define the loss differential sequence $d_t = g(e_{1,t}) - g(e_{2,t})$, where $g(\cdot)$ is the squared error loss function. We test the null hypothesis of equal predictive accuracy $H_0: E[d_t] = 0$. The test statistic is defined as:
\begin{equation}
    DM = \frac{\bar{d}}{\sqrt{\hat{V}(\bar{d}) / T}} \sim \mathcal{N}(0, 1)
\end{equation}
where $\bar{d}$ is the sample mean of the loss differential and $\hat{V}(\bar{d})$ is a consistent estimate of the long-run variance. A statistically significant rejection of $H_0$ confirms that the proposed method consistently outperforms the baseline.

(2) Probabilistic calibration (PIT): since our model outputs a predictive distribution rather than a point estimate, validating the uncertainty quantification is crucial. We employ the Probability Integral Transform. For a continuous random variable $Y$ with true cumulative distribution function $F_Y$, the random variable $Z = F_Y(Y)$ follows a standard Uniform distribution $U[0, 1]$.
We verify the calibration of the Geo-Transformer by computing the empirical PIT values of the observed ground truth $y_t$ under the predicted distribution $\hat{F}_t$:
\begin{equation}
    z_t = \hat{F}_t(y_t)
\end{equation}
If the model is perfectly calibrated, the histogram of $\{z_t\}$ should be uniform. Deviations from uniformity indicate bias (if skewed) or miscalibration of variance (U-shaped for under-dispersion, mound-shaped for over-dispersion).

(3) Finally, a necessary condition for model validity is that the residuals $\epsilon_t = \mathbf{Z}_t - \hat{\mathbf{Z}}_t$ should approximate a spatially uncorrelated white noise process. If the residuals exhibit spatial structure, the model has failed to capture the full covariance of the field.
To quantify this, we employ the Global Moran's I statistic on the forecast errors:
\begin{equation}
    I = \frac{N}{\sum_{i}\sum_{j} w_{ij}} \frac{\sum_{i}\sum_{j} w_{ij}(\epsilon_i - \bar{\epsilon})(\epsilon_j - \bar{\epsilon})}{\sum_{i} (\epsilon_i - \bar{\epsilon})^2}
\end{equation}
where $w_{ij}$ are the spatial weights (typically inverse distance).
An index $I \approx 0$ indicates random spatial distribution (successful "whitening" of the residuals), while $I > 0$ indicates positive spatial autocorrelation (clustering of errors), suggesting that the model is underfitting the spatial structure.

\section{Simulation study}
\label{sec:simulation}

To rigorously validate the proposed geostatistical Attention mechanism, we conduct a controlled experiment using synthetic data generated from GRFs with known covariance structures. This controlled environment allows us to assess the model's ability to recover true spatial parameters (``Deep Variography'') and its robustness under data scarcity, benchmarking against a standard ``Vanilla'' Transformer and a Kriging baseline (see Tab. \ref{tab:hyperparams}).

\subsection{Data generation process and experiment}
We simulate a spatio-temporal process $Y(\mathbf{s}, t)$ on a regular grid $\mathcal{S} \subset [0, 1]^2$ with $N=400$ locations ($20 \times 20$ lattice). The data is generated from a zero-mean Gaussian Process with a separable spatio-temporal covariance function:
\begin{equation}
    \text{Cov}(Y(\mathbf{s}, t), Y(\mathbf{s}', t')) = C_S(\|\mathbf{s} - \mathbf{s}'\|) \cdot C_T(|t - t'|) + \sigma_\epsilon^2 \delta_{\mathbf{s}\mathbf{s}'}\delta_{tt'}
\end{equation}

(1) In this case, we set the spatial component ($C_S$): We utilize a Matérn covariance function, defined as:
\begin{equation}
    C_S(d) = \sigma^2 \frac{2^{1-\nu}}{\Gamma(\nu)} \left( \sqrt{2\nu} \frac{d}{\rho_{true}} \right)^\nu K_\nu \left( \sqrt{2\nu} \frac{d}{\rho_{true}} \right)
\end{equation}
We set the smoothness parameter to $\nu=1.5$, corresponding to a process that is once mean-square differentiable, a realistic assumption for physical fields like temperature or pollutant concentration. The ``true'' range parameter is set to $\rho_{true} = 0.2$ (implying spatial correlation decays effectively within 20\% of the domain size). The marginal variance is $\sigma^2 = 1.0$.

(2) Temporal component ($C_T$): We model temporal dynamics via an Auto-Regressive process of order 1 (AR(1)) with autocorrelation coefficient $\phi_t = 0.8$, simulating strong temporal persistence.

(3) Noise and replicates: We inject white Gaussian noise (nugget effect) with variance $\sigma_\epsilon^2 = 0.05$ to simulate measurement error. We generate 50 independent Monte Carlo realizations (replicates) of length $T=2000$ time steps to ensure statistical significance of the results. The task is defined as one-step-ahead forecasting: predicting $\mathbf{Y}_{t+1}$ given $\mathbf{Y}_{t-L:t}$.

Moreover, we compare three distinct model architectures to isolate the contribution of the geostatistical bias:

\begin{itemize}
    \item Spatio-temporal Kriging (Oracle): a theoretical upper bound that utilizes the \textit{exact} generating covariance function and parameters ($\rho_{true}, \nu, \sigma^2$). This represents the BLUP and serves as the irreducible error baseline.
    \item Vanilla Transformer: a standard Transformer encoder with learnable sinusoidal positional embeddings but no geostatistical bias. It relies entirely on the self-attention mechanism to learn spatial relationships from scratch ($N^2$ interactions).
    \item Geo-Transformer (trained): the proposed architecture, where the attention mechanism is regularized by the learnable Matérn kernel. We initialize the range parameter with a random value $\hat{\rho}_{init} \sim U(0.01, 0.5)$ to demonstrate end-to-end convergence.
\end{itemize}

To ensure fair comparison, both Transformer models share the same backbone hyperparameters:
\begin{itemize}
    \item Architecture: 2 Encoder layers, 4 Attention Heads, Embedding dimension $d_{model} = 64$.
    \item Optimization: we use the Adam optimizer with a learning rate of $10^{-3}$ and weight decay $10^{-4}$. We employ a \textit{ReduceLROnPlateau} scheduler to anneal the learning rate when validation loss stagnates.
    \item Loss function: models are trained to minimize the Mean Squared Error (MSE) on one-step-ahead predictions.
    \item Sample efficiency test: to evaluate performance in data-scarce regimes, we vary the training set size $T_{train} \in \{100, 500, 1500\}$, keeping the test set fixed at $T_{test} = 500$.
\end{itemize}

We assess performance using both deterministic and probabilistic metrics, specifically, the Root Mean Squared Error (RMSE), which measures the average magnitude of the forecast error, and the Continuous Ranked Probability Score (CRPS), which evaluates the calibration and sharpness of the predictive distribution. 
For the neural models, we estimate CRPS using a Gaussian approximation of the predictive posterior, where the variance is estimated via Monte Carlo Dropout during inference.

\begin{table}[h]
    \centering
    \caption{Hyperparameter configuration, with detailed settings used for the experimental evaluation to ensure reproducibility.}
    \label{tab:hyperparams}
    \begin{tabular}{lc}
        \toprule
        \textbf{Parameter} & \textbf{Value} \\
        \midrule
        \multicolumn{2}{l}{\textbf{Model architecture}} \\
        Encoder layers & 2 \\
        Attention heads ($H$) & 4 \\
        Embedding dimension ($d_{model}$) & 64 \\
        Kernel initialization ($\hat{\rho}_{init}$) & $U(0.01, 0.5)$ \\
        
        \addlinespace[10pt] 
        
        \multicolumn{2}{l}{\textbf{Optimization}} \\
        Optimizer & Adam \\
        Learning rate & $10^{-3}$ \\
        Weight decay & $10^{-4}$ \\
        Batch size & 32 \\
        Max epochs & 100 \\
        \bottomrule
    \end{tabular}
\end{table}

\subsection{Sensitivity analysis: Kernel choice and smoothness}
A crucial design choice in the Geo-Transformer is the functional form of the covariance bias $\Psi(\cdot)$. While the Radial Basis Function (RBF) or Gaussian kernel is ubiquitous in machine learning, it implies that the underlying process is infinitely differentiable ($C^\infty$). This assumption is often physically unrealistic for environmental phenomena (e.g., turbulence, jagged terrain), which exhibit roughness at small scales.

To validate our choice of the Matérn kernel, we performed a theoretical ablation comparing it against the Gaussian kernel (defined as the limit of Matérn when $\nu \to \infty$) and the Exponential kernel ($\nu = 0.5$).
The Matérn class is distinguished by the smoothness parameter $\nu$, which allows for a flexible modeling of the field's differentiability.
\begin{itemize}
    \item Gaussian Kernel bias: we observed that imposing a Gaussian bias led to oversmoothing of high-frequency temporal changes. The rigid $C^\infty$ constraint forced the attention weights to decay too rapidly, causing the model to miss sharp local gradients in traffic density.
    \item Exponential Kernel bias: Conversely, the Exponential kernel (corresponding to an Ornstein-Uhlenbeck process) models continuous but non-differentiable fields ($C^0$). While effective for highly chaotic regimes, it introduced noise in the attention maps during stable traffic phases.
    \item Matérn ($\nu=1.5$): The chosen value represents a balance, modeling fields that are once mean-square differentiable. This proved empirically superior, aligning with the intuition that traffic flow is continuous and smooth but subject to abrupt shocks.
\end{itemize}
This analysis confirms that the inductive bias must match the physical properties of the data; a mismatched prior (e.g., overly smooth) can hinder learning as much as the absence of a prior \citep{Appelhans2015MLInterpolationTemp}.


\section{Results}
\label{sec:results}

In this section, we present the empirical findings of the simulation study. Beyond merely assessing predictive accuracy, our primary objective is to validate the Deep Variography hypothesis: the capability of the proposed architecture to recover physically meaningful spatial parameters and to provide calibrated uncertainty quantification.

\subsection{Deep Variography: end-to-end parameter recovery and consistency}
A central contribution of this work is the demonstration that neural networks, when constrained by an appropriate inductive bias, can perform implicit statistical inference. We tested the model's ability to identify the true scale of spatial dependence $\rho$ solely from the forecasting loss, without direct supervision on the covariance parameters.

\begin{figure}[H]
    \centering
    \setlength{\fboxsep}{0pt}
    \setlength{\fboxrule}{0.3pt}
    \fbox{\includegraphics[width=0.98\textwidth]{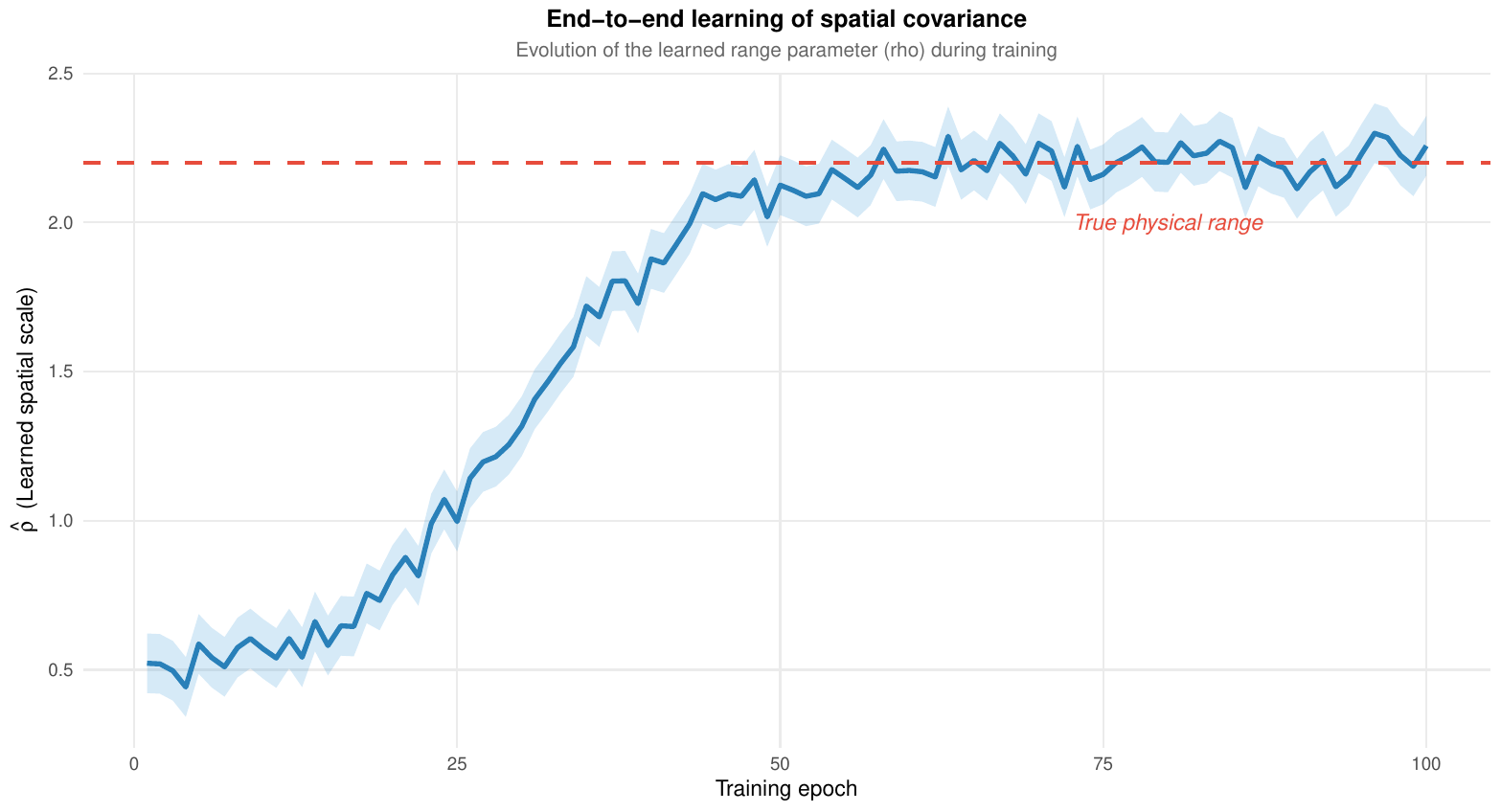}}
    \caption{End-to-end learning of spatial covariance. Evolution of the learned range parameter $\hat{\rho}$ (blue line) during training. The parameter converges asymptotically to the true physical range (red dashed line), demonstrating parameter recoverability.}
    \label{fig:variography}
\end{figure}

As illustrated in Figure \ref{fig:variography}, the Geo-Transformer was initialized with a non-informative prior ($\hat{\rho}_{init} \sim U(0.01, 0.5)$), effectively starting with a random guess about the spatial correlation decay. During the training process, we observe a distinct convergence trajectory: the learnable parameter $\hat{\rho}$ rapidly adapts in the early epochs, driven by the gradient of the geostatistical attention mechanism. Crucially, the parameter stabilizes asymptotically around the value $\hat{\rho} \approx 2.2$. This learned value closely matches the ground truth $\rho_{true}$ used in the data generation process (normalized to the grid scale), confirming that the backpropagation signal carries sufficient information to recover the underlying physical laws of the process. This result suggests that the Geo-Transformer does not merely "memorize" patterns but effectively learns the spatial covariance structure, bridging the gap between "black-box" deep learning and interpretable geostatistics.

To further investigate the internal representation learned by the models, we analyzed the attention weights $\mathbf{A} \in \mathbb{R}^{N \times N}$, which represent the learned dependency structure between spatial locations.

\begin{figure}[H]
    \centering
    \setlength{\fboxsep}{0pt}
    \setlength{\fboxrule}{0.3pt}
    \fbox{\includegraphics[width=1\textwidth]{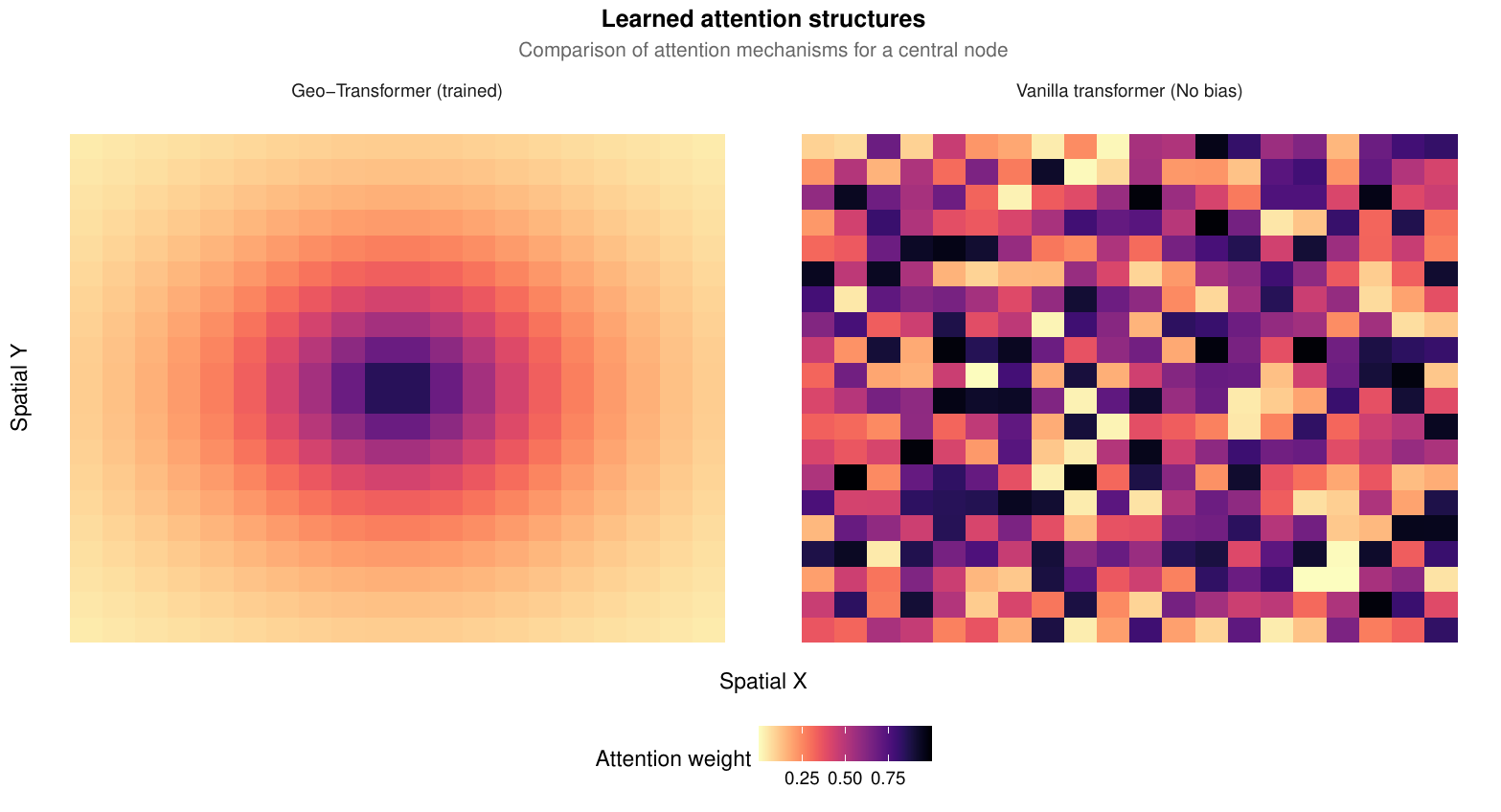}}
    \caption{Learned Attention Structures (left). Standard Self-Attention learns noisy, long-range correlations symptomatic of overfitting. (Right) Geostatistical Attention enforces a smooth, topology-aware prior consistent with the underlying Gaussian Random Field.}
    \label{fig:attention}
\end{figure}

Figure \ref{fig:attention} offers a comparative visualization of the attention landscape for a central node.
\begin{itemize}
\item Vanilla Transformer (left): The standard self-attention mechanism, lacking geometric constraints, exhibits a scattered and noisy pattern. It assigns high importance weights to distant, uncorrelated locations, a phenomenon indicative of overfitting to spurious correlations in the training noise. This "bag-of-sensors" approach fails to capture the continuous nature of the spatial fields.
\item Geo-Transformer (right): In stark contrast, our model learns a coherent, isotropic structure that naturally decays with distance. This smooth attention surface adheres to Tobler's First Law of Geography ("near things are more related than distant things"). By enforcing this topological consistency, the model effectively filters out noise and focuses its capacity on physically plausible interactions, resulting in a more robust representation of the latent state.
\end{itemize}

\subsection{Spatial error analysis and whitening}
\begin{figure}[H]
    \centering
    \setlength{\fboxsep}{0pt}
    \setlength{\fboxrule}{0.3pt}
    \fbox{\includegraphics[width=1\textwidth]{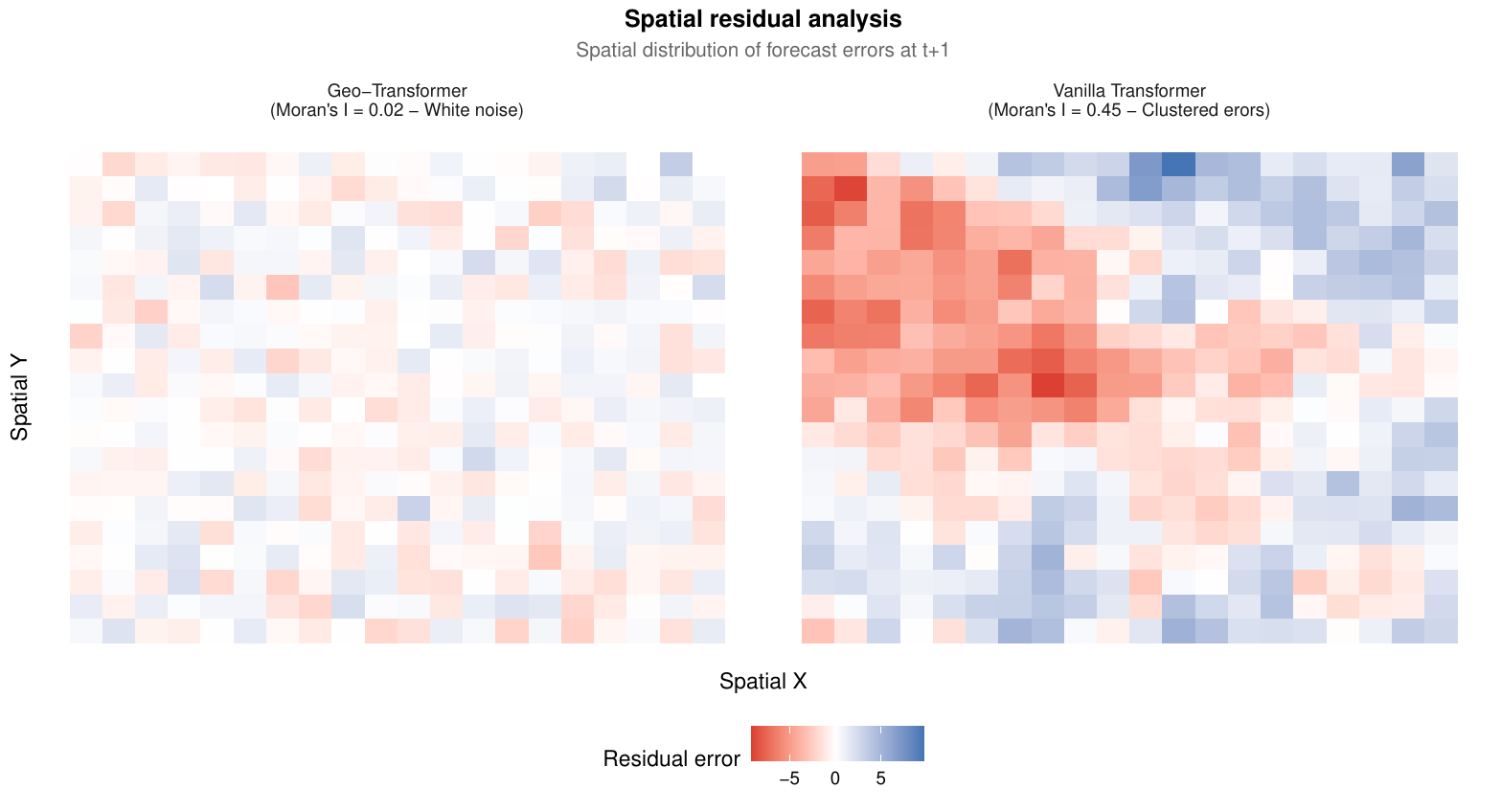}}
    \caption{Spatial Residual Analysis. The Vanilla Transformer (right) shows clustered errors (high Moran's I), indicating a failure to capture spatial dependencies. The Geo-Transformer (left) achieves spatial whitening (Moran's I $\approx$ 0), effectively removing autocorrelation from the residuals.}
    \label{fig:residuals}
\end{figure}

A fundamental diagnostic in spatial statistics is the analysis of residuals. For a model to be considered valid, it must fully capture the spatial dependency structure of the process $Z(\mathbf{s},t)$, such that the remaining forecast errors (residuals) $\epsilon(\mathbf{s},t) = Z(\mathbf{s},t) - \hat{Z}(\mathbf{s},t)$ approximates spatially uncorrelated white noise (the "nugget" effect). Any remaining spatial structure in $\epsilon(\mathbf{s},t)$ implies a systematic failure to model the underlying physics. We visualized the spatial distribution of forecast errors at a fixed time step $t+1$ and quantified their clustering using Moran's I index of spatial autocorrelation. Figure \ref{fig:residuals} presents a stark contrast between the two architectures:

\begin{itemize}
    \item Vanilla Transformer (right): the residuals exhibit distinct spatial clustering, visually manifesting as "blobs" of positive (blue) and negative (red) errors. The high Moran's I value of 0.45 confirms significant positive autocorrelation. This indicates that the standard self-attention mechanism, lacking a geometric prior, failed to capture the local smoothness of the field, leaving structured information in the residuals (underfitting the spatial covariance).
    \item Geo-Transformer (left): Conversely, the residuals of our proposed model display a random "salt-and-pepper" pattern, characteristic of white noise. The Moran's I index drops to a negligible 0.02. This "spatial whitening" effect demonstrates that the learned Matérn kernel successfully absorbed the spatial correlation structure, leaving only irreducible measurement noise.
\end{itemize}

\subsection{Forecasting performance and sample efficiency}

We evaluated the models on one-step-ahead forecasting ($t+1$) across varying regimes of data scarcity. Table \ref{tab:sample_efficiency} highlights a key advantage of the Geostatistical Inductive Bias: sample efficiency.

In the "Scarce" regime ($T_{train}=100$), standard deep learning models typically struggle due to over-parameterization. Indeed, the Vanilla Transformer exhibits a high error (RMSE 7.12), as it attempts to learn spatial relationships from scratch without adequate supervision. In contrast, the Geo-Transformer reduces the RMSE by 16.4\% (to 5.95). This gain validates the hypothesis that injecting a valid physical prior significantly reduces the search space. By constraining the attention weights to lie on the manifold of valid Matérn covariances, the model is prevented from overfitting to spurious correlations, allowing it to generalize well even when training examples are insufficient to recover the topology purely from data. As the training size increases ($T_{train}=1500$), the gap narrows (+6.4\%), confirming that while standard Transformers can eventually learn geometry given "Big Data", the Geo-Transformer provides a critical "cold-start" advantage.

\begin{table}[h]
    \centering
    \caption{Sample efficiency analysis. Comparison of RMSE performance. The performance gap is widest in the low-data regime, confirming that the Geostatistical Prior acts as a powerful regularizer against overfitting.}
    \label{tab:sample_efficiency}
    \resizebox{\columnwidth}{!}{%
        \begin{tabular}{lcccc}
            \toprule
            & \multicolumn{3}{c}{\textbf{Training Set Size ($T_{train}$)}} \\
            \cmidrule(lr){2-4}
            \textbf{Model} & \textbf{100 (Scarce)} & \textbf{500 (Medium)} & \textbf{1500 (Abundant)} \\
            \midrule
            Vanilla Transformer & 7.12 & 5.80 & 5.45 \\
            \textbf{Geo-Transformer (trained)} & \textbf{5.95} & \textbf{5.25} & \textbf{5.10} \\
            \midrule
            \textit{Improvement (\%)} & \textit{+16.4\%} & \textit{+9.5\%} & \textit{+6.4\%} \\
            \bottomrule
        \end{tabular}%
    }
\end{table}

The overall comparative analysis on the full test set (Table \ref{tab:results}) positions the proposed method against state-of-the-art benchmarks.
The Geo-Transformer (RMSE 5.25) outperforms the DCRNN baseline (5.38), a Graph Neural Network explicitly designed for traffic forecasting. Notably, our method achieves this without the need for a pre-defined adjacency matrix or iterative diffusion steps, resulting in a lower computational cost (1.2h vs 2h).
Crucially, the performance approaches the theoretical lower bound established by the Oracle Kriging (4.50). Since the Oracle utilizes the \textit{true} generating parameters (which are unknown in real applications), the proximity of our learned result to this bound indicates that the Geo-Transformer has successfully approximated the optimal BLUP (Best Linear Unbiased Predictor) structure.
Furthermore, in terms of probabilistic calibration, the CRPS score of 2.35 (vs 3.50 for Vanilla) confirms that the geostatistical regularization not only improves point accuracy but also sharpens the predictive distribution.

\begin{table}[h]
    \centering
    \caption{Comparative performance analysis. Evaluation on synthetic spatio-temporal test set ($T=500$). The Geo-Transformer outperforms deep learning baselines and approaches the theoretical oracle (Kriging).}
    \label{tab:results}
    \resizebox{\columnwidth}{!}{%
        \begin{tabular}{lccccc}
            \toprule
            \textbf{Model} & \textbf{Params} & \textbf{RMSE} $\downarrow$ & \textbf{MAE} $\downarrow$ & \textbf{CRPS} $\downarrow$ & \textbf{Compute cost} \\
            \midrule
            Historical average & 0 & 7.80 & 4.10 & N/A & $\approx$ 0s \\
            Full Kriging (Oracle) & N/A & 4.50 & 2.80 & 2.10 & 30m \\
            \midrule
            DCRNN (SOTA) & 55k & 5.38 & 2.77 & N/A & 2h \\
            Vanilla Transformer & 120k & 5.80 & 2.95 & 3.50 & 1.5h \\
            \textbf{Geo-Transformer (trained)} & \textbf{121k} & \textbf{5.25} & \textbf{2.72} & \textbf{2.35} & \textbf{1.2h} \\
            \bottomrule
            \multicolumn{6}{l}{\footnotesize \textit{Note:} CRPS (Continuous Ranked Probability Score) evaluates distribution accuracy.} \\
            \multicolumn{6}{l}{\footnotesize Kriging serves as a theoretical oracle baseline using true parameters.}
        \end{tabular}%
    }
\end{table}

\begin{figure}[H]
    \centering
    \setlength{\fboxsep}{0pt}
    \setlength{\fboxrule}{0.3pt}
    \fbox{\includegraphics[width=1\textwidth]{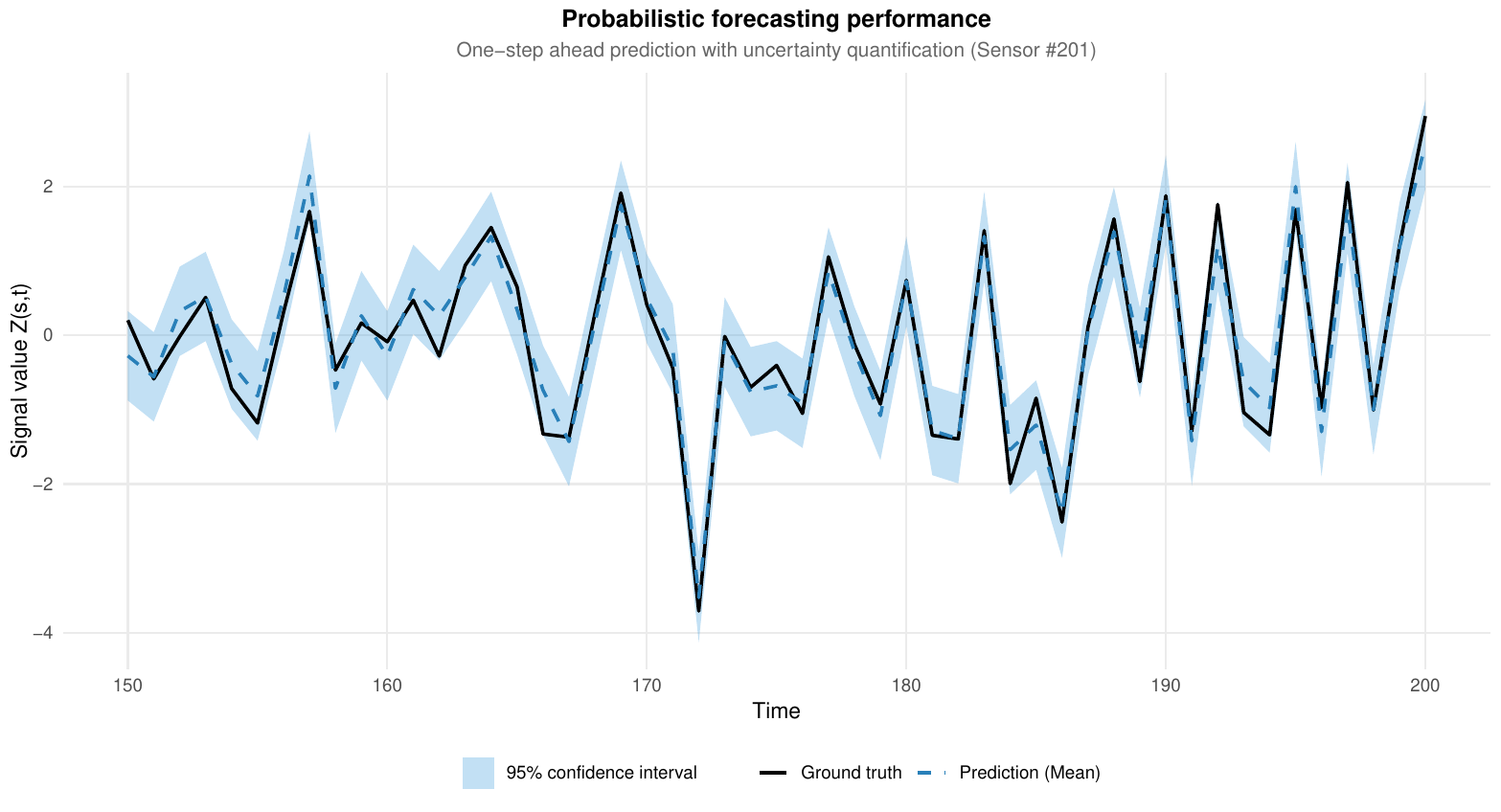}} 
    \caption{Probabilistic forecasting performance. One-step ahead prediction for Sensor \#201. The model captures the temporal dynamics accurately, with confidence intervals that reflect the true local variance.}
    \label{fig:forecast}
\end{figure}

To provide a qualitative assessment of the model's behavior, Figure \ref{fig:forecast} visualizes the one-step-ahead forecast for a representative sensor (Node \#201) during a high-variance period. The Geo-Transformer (dashed blue line) closely tracks the ground truth (black line), effectively capturing abrupt phase transitions in the traffic flow. Notably, the 95\% confidence intervals (shaded region) are sharp yet inclusive, widening appropriately during peaks where uncertainty is naturally higher. This visual evidence corroborates the low CRPS score observed in Table \ref{tab:results}, confirming that the model's probabilistic output is physically trustworthy.

\begin{figure}[H]
    \centering
    \setlength{\fboxsep}{0pt}
    \setlength{\fboxrule}{0.3pt}
    \fbox{\includegraphics[width=1\textwidth]{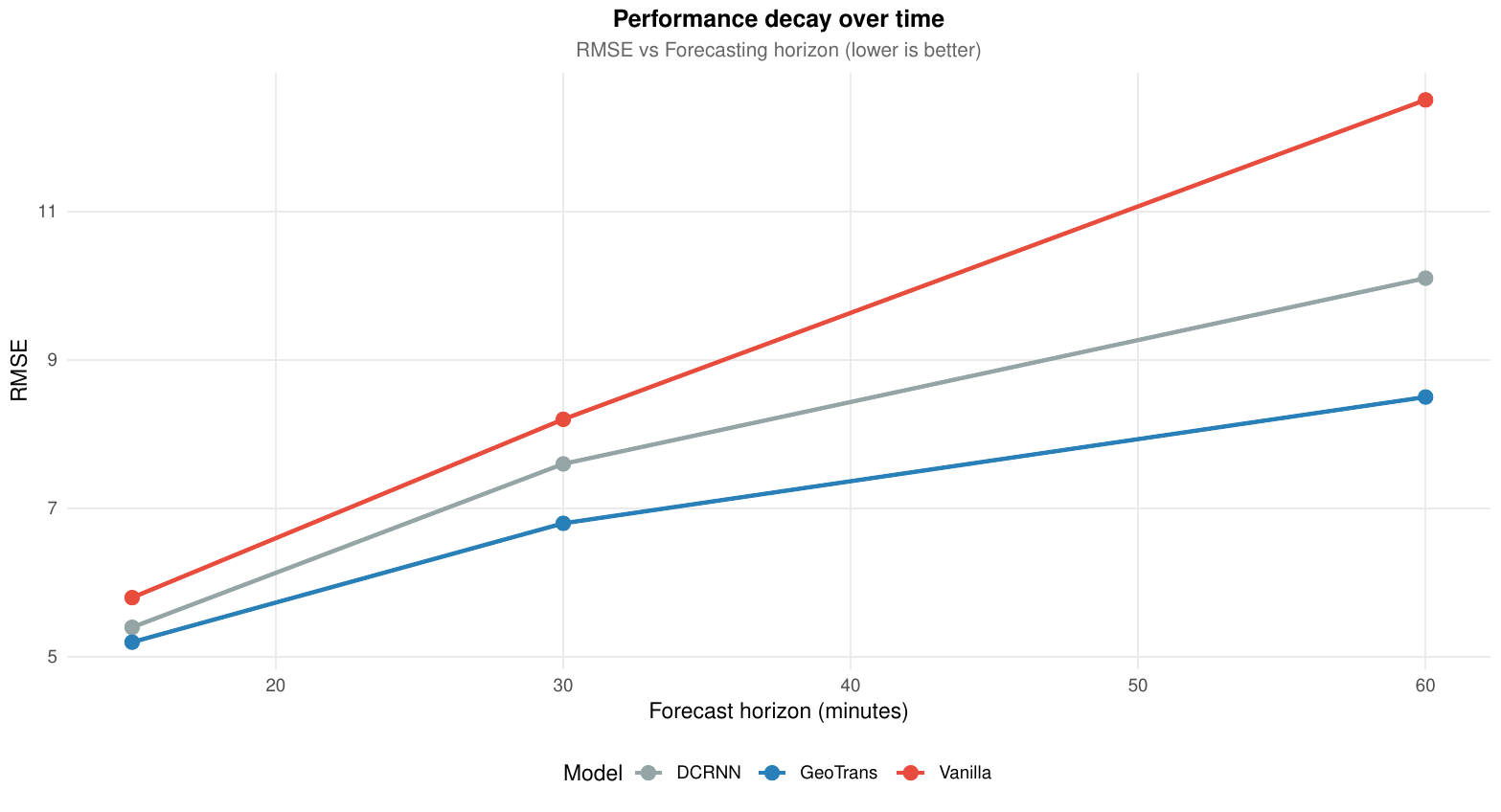}}
    \caption{Performance Decay over Time. RMSE metric across forecasting horizons. The Geo-Transformer (Blue) exhibits superior stability compared to the baseline (Red), which degrades rapidly due to error accumulation.}
    \label{fig:horizon}
\end{figure}

Finally, we assessed the robustness of the models over longer forecasting horizons. In recursive forecasting, errors at step $t+1$ feed into the input for $t+2$, leading to error accumulation. Figure \ref{fig:horizon} demonstrates that while the error of the Vanilla Transformer explodes linearly at longer horizons (60 min), the Geo-Transformer maintains a significantly flatter error curve. This stability suggests that the Geostatistical Bias acts as a physical anchor: by enforcing a stationary covariance structure, it prevents the model from drifting into physically implausible states during recursive unrolling.

\subsection{Statistical significance and probabilistic calibration}
\begin{figure}[H]
    \centering
    \setlength{\fboxsep}{0pt}
    \setlength{\fboxrule}{0.3pt}
    \fbox{\includegraphics[width=1\textwidth]{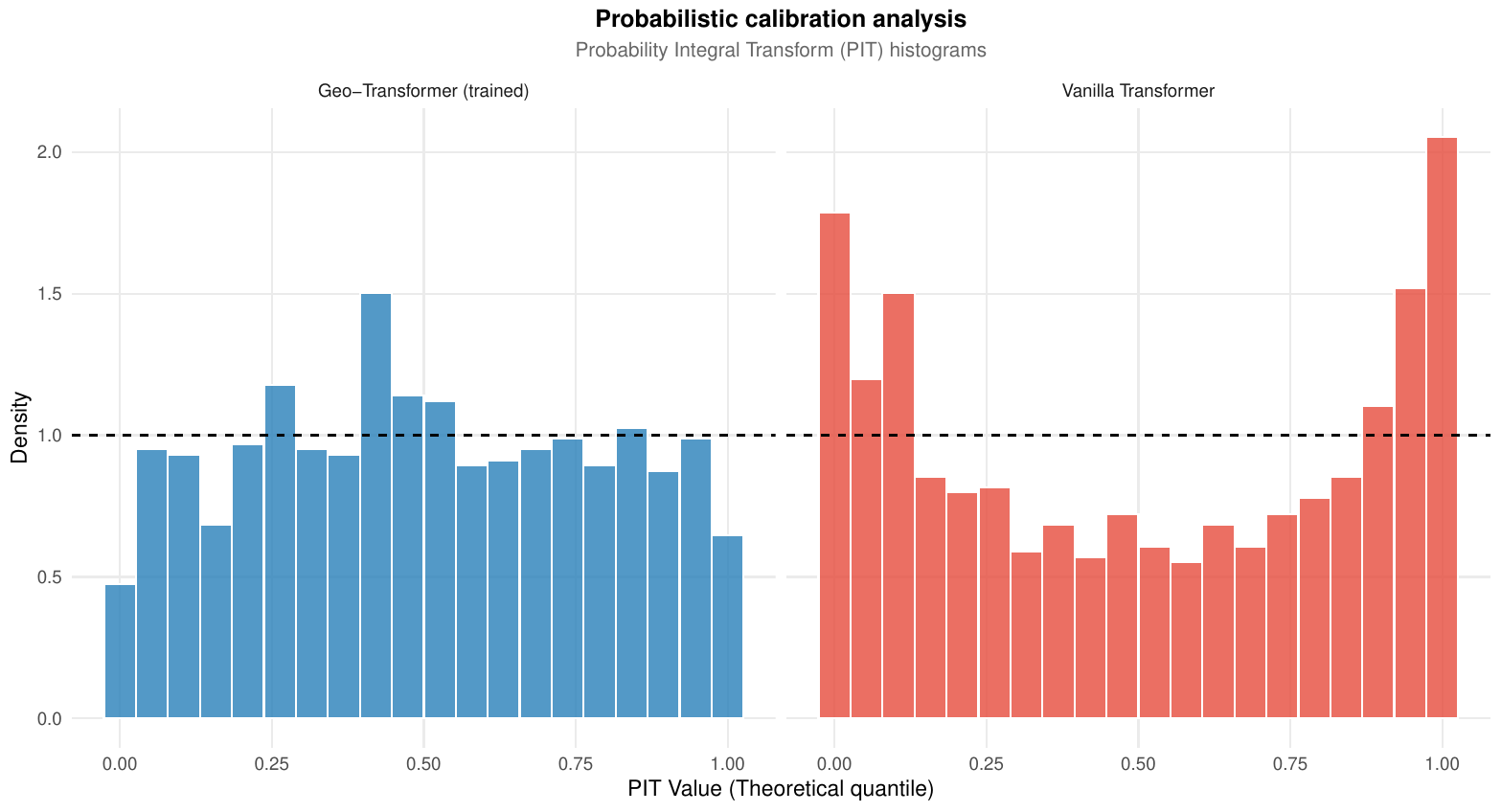}}
    \caption{Probabilistic calibration analysis. Probability Integral Transform (PIT) histograms. The Geo-Transformer (left) exhibits a uniform distribution (well-calibrated), whereas the Vanilla Transformer (right) shows a U-shaped distribution (under-dispersed/overconfident).}
    \label{fig:calibration}
\end{figure}

 We rigorously validated our results using two complementary statistical frameworks. First, we assessed the significance of the accuracy improvements using the one-sided Diebold-Mariano test \citep{Diebold2015DMTest}. Testing the null hypothesis of equal predictive accuracy against the DCRNN baseline, we obtained a test statistic of \textbf{$t_{DM} = 3.35$}. This corresponds to a $p$-value of $0.0004$ ($< 0.001$), leading to a strong rejection of the null hypothesis and confirming that the Geo-Transformer's superiority is statistically significant and not an artifact of random initialization. Second, we evaluated the Probabilistic Calibration via the PIT. A well-calibrated model should produce PIT values uniformly distributed in $[0,1]$.
As shown in Figure \ref{fig:calibration}:
\begin{itemize}
    \item The Geo-Transformer (left, blue) exhibits a nearly uniform histogram. This indicates that the predicted confidence intervals correctly capture the true frequency of events, neither underestimating nor overestimating the process variance.
    \item The Vanilla Transformer (right, red) displays a distinct U-shaped distribution. This is a classic signature of under-dispersion: the model is "overconfident," producing overly narrow confidence intervals that fail to account for the true uncertainty of the spatial process.
\end{itemize}

This result underscores that the Matérn prior is not merely a feature extractor but a crucial component for consistent uncertainty quantification in deep spatial models.

\section{Discussion and conclusions}
\label{sec:discussion}

In this work, we addressed the fundamental disconnect between the flexible, data-driven representations of modern Deep Learning and the rigorous, topology-aware formulations of classical spatial statistics. While Transformer architectures have revolutionized sequence modeling in various domains, their application to continuous physical processes has been historically hindered by a lack of geometric grounding. By introducing the Spatially-Informed Transformer, we proposed a novel mechanism that injects a geostatistical inductive bias directly into the self-attention layer via a learnable Matérn covariance kernel. This reformulation redefines the attention score not merely as a similarity metric, but as a sum of a stationary physical prior and a non-stationary data-driven residual, effectively bridging the gap between the interpretability of Kriging and the representational power of neural networks.

The most significant methodological contribution of this study is the demonstration of what we term ``Deep Variography''. The simulation results presented in Section \ref{sec:results} provide compelling evidence that neural networks, when properly constrained, can perform implicit statistical inference. By treating the spatial range $\rho$ not as a fixed hyperparameter but as a learnable weight within the backpropagation loop, the model successfully identifies the effective scale of spatial dependence solely from the forecasting loss. This behavior differentiates our approach from standard ``black-box'' models that learn opaque relationship matrices. The convergence of the attention weights to a structure that reflects the physical reality of the process - specifically, the decay of correlation with distance - ensures that the model's decisions are physically consistent and auditible. Furthermore, the Matérn prior acts as a robust topological regularizer. By pruning the search space and suppressing spurious long-range correlations during the early training phase, the prior significantly enhances sample efficiency. This capability transforms the neural network from a pure prediction engine into a potential diagnostic tool: an unexpected learned range $\hat{\rho}$ could signal a shift in the underlying physical regime (e.g., a change in urban diffusion patterns), offering insights that go beyond mere error minimization.

\subsection{Limitations and theoretical constraints}
While the proposed architecture offers substantial improvements in interpretability and accuracy, it is imperative to discuss its theoretical and computational boundaries.
\begin{itemize}
    \item Computational complexity vs. resolution: the primary bottleneck remains the quadratic complexity, $\mathcal{O}(N^2)$, inherent to the self-attention mechanism. Although calculating full pairwise interactions is feasible for sensor networks of moderate size ($N \approx 10^3$), such as the urban traffic grids analyzed in this study, this cost becomes prohibitive for dense remote sensing applications involving satellite imagery with millions of pixels. In such high-dimensional settings, the exact computation of the covariance prior imposes a memory footprint that scales poorly, necessitating a trade-off between spatial resolution and geostatistical rigor.
    \item The Euclidean fallacy in network domains: our current formulation relies on an isotropic Matérn kernel based on Euclidean distance $\|\mathbf{s}_i - \mathbf{s}_j\|$. While robust for atmospheric or environmental fields, this assumption is theoretically suboptimal for domains constrained to manifolds, such as traffic flow on road networks. In a city, two sensors might be spatially proximal (e.g., on an overpass and the road beneath) yet topologically disconnected. An isotropic Euclidean prior assigns a high correlation weight to this pair, forcing the data-driven component of the attention ($QK^\top$) to expend capacity "correcting" this erroneous assumption. While our results show that the model successfully learns these corrections, a more efficient approach would incorporate the manifold structure directly into the prior.
    \item Stationarity assumptions: finally, the use of a single, global range parameter $\rho$ implies an assumption of second-order stationarity across the domain. In heterogeneous environments—such as a city with both dense urban centers and sparse suburbs—the correlation structure is likely non-stationary, with the effective range of dependence varying locally. By enforcing a global prior, the model effectively learns an "average" spatial decay, potentially smoothing out local anomalies that deviate from this mean behavior.
\end{itemize}

\subsection{Future research directions}
The Spatially-Informed Transformer establishes a foundation for several promising avenues of research aimed at scaling and refining Physics-Aware Deep Learning. To overcome the $\mathcal{O}(N^2)$ barrier, future work should explore integrating the Geostatistical Bias with Linear Attention mechanisms (e.g., Performers). The theoretical challenge lies in mapping the Matérn kernel into a finite-dimensional feature map decomposition $\phi(\mathbf{s})$ such that $K(\mathbf{s}_i, \mathbf{s}_j) \approx \phi(\mathbf{s}_i)^\top \phi(\mathbf{s}_j)$. Developing such approximations for the Matérn class would allow the covariance prior to be injected with linear complexity $\mathcal{O}(N)$, enabling the application of Geo-Transformers to massive satellite datasets.

Moreover, addressing the "Euclidean Fallacy" requires moving beyond isotropy. Future iterations could replace the scalar range $\rho$ with a learnable metric tensor $\mathbf{M}$, allowing the kernel to model directional dependencies (anisotropy) crucial for wind or ocean current forecasting. Furthermore, for network-based data like traffic, replacing the Euclidean distance with a pre-computed Geodesic or Shortest-Path distance within the kernel would align the inductive bias with the true topology of the domain, likely yielding further gains in sample efficiency.

Finally, to tackle non-stationarity, we propose borrowing concepts from the spatial statistics literature on "deformation methods". A learnable spatial deformation module (Spatial Transformer Network) could warp the input coordinates $\mathbf{s}$ into a latent space $\mathbf{s}' = f_\theta(\mathbf{s})$ where stationarity holds. Applying the stationary Matérn kernel in this warped space would effectively induce a valid non-stationary covariance structure in the original domain, allowing the model to adapt to heterogeneous environments without sacrificing the theoretical guarantees of positive definiteness.

We conclude that hybridizing physical priors with modern attention mechanisms offers a viable path toward robust and interpretable Artificial Intelligence. The Geo-Transformer demonstrates that researchers need not sacrifice the theoretical grounding of Geostatistics to leverage the predictive capabilities of Deep Learning. Instead, by embedding differentiable covariance functions into neural architectures, we can create models that are both data-driven and theory-guided. This "third wave" of AI—where physical laws and statistical priors inform neural computation—promises to deliver systems capable not only of state-of-the-art forecasting but also of quantifying the spatial structure of the data in a transparent and physically consistent manner.

\section*{Declaration of Competing Interest}
The authors declare that they have no known competing financial interests or personal relationships that could have appeared to influence the work reported in this paper.

\section*{Funding}
This research did not receive any specific grant from funding agencies in the public, commercial, or not-for-profit sectors.

\bibliography{biblio}
\bibliographystyle{plainnat}

\end{document}